\documentclass[10pt,letterpaper]{article}

\usepackage[left=1in,right=1in,top=1in,bottom=1in]{geometry}
\usepackage[T1]{fontenc}
\usepackage[utf8]{inputenc}
\usepackage{lmodern}
\usepackage{microtype}
\usepackage{url}
\ifdefined\pdfgentounicode\pdfgentounicode=1\fi

\usepackage{amsmath,amssymb}
\usepackage{graphicx}
\usepackage{cite}
\usepackage{indentfirst}
\usepackage{authblk}
\makeatletter
\renewcommand\AB@affilnote[1]{\raisebox{0.35ex}{\scriptsize#1}\hspace{0.08em}}
\makeatother
\usepackage[font=small,labelfont=bf]{caption}
\usepackage[hidelinks]{hyperref}
\hypersetup{
    pdftitle={SAT-Edge-Agent: Hardware-in-the-Loop Edge-Agent Orchestration for Onboard Satellite Intelligence},
    pdfauthor={Longji He and Jeto Xu},
    pdfsubject={Hardware-in-the-loop satellite edge-agent orchestration},
    pdfkeywords={satellite edge computing, hardware-in-the-loop, onboard intelligence, edge agent, COTS edge SoC}
}
\usepackage{titlesec}
\usepackage{enumitem}
\usepackage{float}

\usepackage[table,xcdraw,dvipsnames]{xcolor}
\usepackage{booktabs}
\usepackage{multirow}
\usepackage{pifont}
\usepackage{textcomp}
\usepackage{array}

\definecolor{npsgreen}{HTML}{E1F1EC}
\definecolor{npsred}{HTML}{FEECDD}
\definecolor{cvprblue}{HTML}{DEEBF7}
\definecolor{icmlred}{HTML}{E58579}
\definecolor{cvprgreen}{HTML}{BBDBA6}
\definecolor{cvprorange}{HTML}{F4B183}
\definecolor{lightblue}{HTML}{DCEAF7}
\definecolor{lightgreen}{HTML}{E8F3E1}
\definecolor{lightorange}{HTML}{F9E7D2}
\definecolor{graybg}{gray}{0.95}

\usepackage{tikz}
\usetikzlibrary{shapes.geometric, arrows.meta, positioning, fit, backgrounds, shadows, calc}

\tikzset{
    basebox/.style={rectangle, draw=gray!60, thick, rounded corners=3pt, align=center, fill=white, font=\small, drop shadow={opacity=0.05}},
    frontbox/.style={basebox, fill=cvprblue!45, minimum height=1cm},
    agentbox/.style={basebox, fill=npsgreen!80, minimum height=1cm},
    toolbox/.style={basebox, fill=cvprorange!45, minimum height=1cm},
    edgebox/.style={basebox, fill=lightgreen!75, minimum height=1cm},
    note/.style={basebox, fill=gray!8, font=\scriptsize},
    arrow/.style={->, >=Stealth, thick, draw=gray!80},
    dasharrow/.style={->, >=Stealth, dashed, thick, draw=gray!80}
}

\renewcommand{\thesection}{\Roman{section}}
\titleformat{\section}{\normalfont\bfseries\centering}{\thesection.}{0.6em}{\MakeUppercase}
\titleformat{name=\section,numberless}{\normalfont\bfseries\centering}{}{0pt}{\MakeUppercase}
\titleformat{\subsection}{\normalfont\bfseries}{}{0pt}{}
\titlespacing*{\section}{0pt}{1.25\baselineskip}{0.75\baselineskip}
\titlespacing*{\subsection}{0pt}{0.9\baselineskip}{0.45\baselineskip}
\setlist[enumerate]{leftmargin=1.5em,itemsep=0.25em,topsep=0.35em}

\title{SAT-Edge-Agent: Hardware-in-the-Loop Edge-Agent Orchestration\\
for Onboard Satellite Intelligence}

\author{}
\date{}
\newcommand{\affline}[2]{%
  \noindent\hangindent=1.5em\hangafter=1%
  \makebox[1.5em][r]{\textsuperscript{#1}}#2\par
}

\begin{document}
\maketitle
\vspace{-4.4em}
\begin{center}
{\fontsize{12}{15}\selectfont
Longji He$^{1,2}$ and Jeto Xu$^{1}$}
\end{center}
\vspace{0.55em}
{\fontsize{10}{14}\selectfont
\affline{1}{OgCloud Limited, Guangzhou, GD, China, \texttt{keith@ogcloud.com}}
\affline{2}{College of Engineering, The Pennsylvania State University, University Park, PA, USA}
}
\vspace{0.85em}

\begin{abstract}

Onboard satellite intelligence requires more than model execution: a task layer must translate mission intent into local tool calls, expose execution state, and return machine-consumable artifacts under communication and power constraints. We present SAT-Edge-Agent, a hardware-in-the-loop (HIL) edge-agent system deployed on a commercial off-the-shelf (COTS), ARM-based heterogeneous edge system-on-chip (SoC). A browser workspace and FastAPI agent coordinate a local OpenAI-compatible language service with a project-internal YOLO-style oriented-object-detection endpoint that returns FAIR1M metadata-backed structured results. We repeated two fixed FAIR1M workloads 20 times each: one single-image request and one serial two-image request. Both workloads completed 20/20 attempts. Mean Full-Agent latency was 29.353 s and 60.937 s, with nearest-rank empirical P95 values of 31.166 s and 66.882 s. Mean cumulative detector time was 861.386 ms and 1510.920 ms, only 2.93\% and 2.48\% of the corresponding Full-Agent means. A separate profiler confirms that most visible latency occurs outside detector execution, before and after the first visible response token. Mean CPU utilization was 20.761\% and 20.482\%. A devfreq/sysfs NPU-load field sampled every 200 ms had a mean of 100\% for both workloads; because the detector and local language service share the accelerator, this field is neither detector-only occupancy nor calibrated utilization. The public evidence package provides sanitized request-level records, redacted JSON, normalized success and partial-result SSE examples, and scripts that reproduce the reported statistics. An earlier unsynchronized plug-meter pilot is retained only as board-level context. These results establish a reproducible HIL boundary for observable satellite edge-agent orchestration. They do not establish detector accuracy, a new geolocation method, calibrated energy efficiency, or flight readiness.

\end{abstract}

\section{Introduction}

Low-Earth-orbit Earth-observation missions increasingly require interpretation to move closer to the sensor. A satellite that only stores and forwards raw imagery remains dependent on communication windows, downlink bandwidth, ground-side inference, and manual task interpretation. Onboard AI changes this pattern by allowing the spacecraft or payload computer to filter low-value scenes, identify high-value events, and return compact mission products before the full image archive is downlinked \cite{phisat1,cloudscout,ravaen}. Recent orbital-edge and satellite-edge studies make the same systems point in a broader form: sensing volume, contact intermittency, latency, and energy constraints make purely ground-side processing an increasingly poor default for time-sensitive missions \cite{oec_asplos,satedge_tcom,ai_onboard_survey}.

A detector alone does not provide this operational layer. An onboard-intelligence payload must interpret mission intent, select and invoke local tools, move data between perception and language services, expose intermediate state, and return outputs for both software and operators. This requirement becomes more important as satellite-edge research considers large models and microservice-style inference near the sensor \cite{satellite_edge_large_models}. In this setting, the spacecraft is not only a remote sensor; it is a constrained edge computer that can produce mission artifacts before raw-data downlink decisions are made.

SAT-Edge-Agent addresses this systems problem in a HIL setting. A low-power COTS ARM-based heterogeneous edge-SoC platform hosts local inference services behind an agent workflow. The vision service returns oriented detections and FAIR1M metadata-backed target fields from remote-sensing images \cite{fair1m}. The backend streams the structured tool artifact as the authoritative machine-facing result and uses a local OpenAI-compatible language service to form an optional operator-facing summary. This separation allows downstream software to consume detection results without depending on the natural-language path.

The paper therefore asks two bounded systems questions: can local perception and language services be organized as an observable end-to-end agent workflow on a low-power edge SoC, and where does request latency arise within that workflow? The contribution is neither a new YOLO architecture nor a flight demonstration. A detector study would require an accuracy benchmark, training protocol, and model-level baselines. This HIL study instead evaluates integration, service contracts, execution visibility, fixed-workload repeatability, and a public reproduction boundary constrained by data and model licenses.

Power awareness remains part of the systems boundary, but the current evidence is intentionally secondary. A separate 2026-07-03 plug-meter pilot is retained in \ref{app:artifacts} as board-level context. It was not recorded concurrently with the repeated 2026-07-13 FAIR1M latency experiment and is therefore not used to claim calibrated power, energy per request, or spacecraft energy efficiency.

The paper makes four contributions:
\begin{enumerate}
    \item An end-to-end HIL architecture that combines a browser workspace, FastAPI agent backend, local language service, and project-internal YOLO-style OBB endpoint on a COTS ARM-based heterogeneous edge SoC.
    \item A documented output and reproducibility contract that separates machine-facing tool artifacts from optional narrative output, and public source and service contracts from private weights and laboratory infrastructure.
    \item Structured remote-sensing artifacts containing class labels, confidence scores, oriented polygons, pixel centers, and FAIR1M metadata-backed geographic target fields, without presenting metadata propagation as a new geolocation method.
    \item Repeated fixed-workload measurements of Full-Agent and YOLO-tool latency, visible-response timing, observed completion, and CPU/NPU telemetry for single-image and serial two-image workflows, with the earlier plug-meter pilot kept separate from the repeated experiment.
\end{enumerate}

\section{Related Work}

\subsection{Onboard Autonomy and Satellite AI}

Autonomous spacecraft and onboard sciencecraft systems show that spacecraft can detect events, plan responses, and reduce dependence on continuous ground control. EO-1-style autonomous science operations and onboard classifiers for science-event detection are important precedents because they demonstrate that onboard software can identify high-value observations and alter what is returned to the ground \cite{eo1_sciencecraft,onboard_classifiers}. Recent Earth-observation AI missions such as Phi-Sat-1 and CloudScout show that neural-network inference can be executed onboard satellite payloads for tasks such as cloud filtering and downlink selection \cite{phisat1,cloudscout}. RaVAEn and later onboard-training work further show that compact learned representations can support onboard change detection, inference, and limited training under satellite hardware constraints \cite{ravaen,fast_onboard_training}. These missions motivate SAT-Edge-Agent's direction, but the present paper occupies a different layer. It does not only ask whether a neural model can run onboard; it asks how edge inference can be wrapped inside a mission-facing agent workflow with a reproducible HIL boundary.

\subsection{Satellite Edge Computing and COTS Hardware}

Satellite edge computing treats orbital and near-orbital nodes as compute systems close to the sensor \cite{oec_asplos,satedge_tcom}. This perspective is useful for missions where raw data volume, latency, ground-station visibility, and energy constrain operational value. Recent surveys and workshop reports also emphasize the practical obstacles: restricted power, memory, storage, radiation environment, incomplete datasets, and the need to benchmark AI workloads on realistic embedded processors \cite{ai_onboard_survey,gardill_space_edge}. Broader heterogeneous edge-platform benchmarking further shows that CPU, GPU, and NPU advantages depend on the workload and metric, so processor-level speedup values should not be transferred across unlike tasks \cite{edge_ai_platform_benchmark}. The edge-SoC evidence in this paper should be read in that HIL sense. It supports a ground-based edge-payload prototype and a power-aware software stack, not a radiation-qualified flight-computer claim.

\subsection{Agent Orchestration for Edge AI}

Tool-using LLM systems offer an abstraction for deciding when to call detectors, how to combine tool results, and how to expose intermediate state. ReAct studies interleaved reasoning and action, Toolformer studies learned API use, and recent surveys organize LLM-based agents around planning, memory, tools, and evaluation \cite{react,toolformer,llm_agents_survey}. A recent Earth-observation preprint further explores hierarchical multi-agent routing for onboard crisis-response analysis on an engineering edge platform \cite{onboard_multiagent_eo}. SAT-Edge-Agent adopts this abstraction as an engineering control layer. The local language model is not presented as a complete autonomous mission planner; it converts local tool results into an observable, mission-facing workflow.

\subsection{Remote-Sensing OBB Detection and FAIR1M}

Remote-sensing imagery often contains rotated, dense, and fine-grained targets. DOTA helped establish oriented object detection in aerial imagery by annotating objects with arbitrary quadrilaterals, while FAIR1M provides high-resolution imagery, oriented bounding boxes, fine-grained categories, and geographic metadata suitable for aircraft, ship, and vehicle analysis \cite{dota,fair1m}. Remote-sensing detection reviews and OBB surveys further show why orientation-aware target representations are important in optical remote-sensing imagery \cite{rsod_review,obb_survey}. SAT-Edge-Agent uses this style of data because the output artifacts are naturally mission-oriented: classes, confidence, oriented polygons, and dataset-provided geographic fields can be serialized into structured task results. In this paper, FAIR1M and the YOLO26-labeled service support the agent workflow as a vision tool. They are not used to claim a new state-of-the-art detector or a new geolocation algorithm.

\subsection{Comparative Positioning and System-Level Efficiency}
\label{sec:comparative_positioning}

SAT-Edge-Agent must be positioned at the system layer measured by the experiment. Existing onboard-AI work establishes that neural models can execute close to the sensor: Phi-Sat-1 and CloudScout demonstrate cloud filtering for downlink selection, RaVAEn demonstrates compact representation learning and onboard model execution, and Myriad/Snapdragon studies benchmark model-dependent acceleration on COTS or COTS-derived processors under space-relevant conditions \cite{phisat1,cloudscout,ravaen,fast_onboard_training,edge_processor_benchmark}. Their reported metrics are primarily model-level quantities such as accuracy, encoding or inference time, quantization discrepancy, power, and energy per inference. They are therefore precedents and measurement references, not direct baselines for a workflow that combines a detector endpoint, local language service, streamed state, and structured mission-result generation.

\begin{table}[htbp]
\centering
\small
\renewcommand{\arraystretch}{1.12}
\begin{tabular}{p{0.19\linewidth}p{0.24\linewidth}p{0.24\linewidth}p{0.24\linewidth}}
\toprule
\textbf{Reference layer} & \textbf{Reported value type} & \textbf{What it supports} & \textbf{Comparison boundary} \\
\midrule
Phi-Sat-1 / CloudScout & Onboard cloud-filtering accuracy, embedded inference, and downlink-selection motivation & Onboard AI can remove or deprioritize low-value Earth-observation data before downlink & Different task and output artifact; not a direct latency or accuracy baseline for OBB Agent workflows \\
RaVAEn onboard model execution & Tile-level representation encoding and onboard model execution/training evidence & Compact learned representations can be executed near the sensor and used for downstream prioritization & Different model, data unit, and mission product; useful for context, not a speedup denominator \\
Myriad/Snapdragon benchmark & Processor-dependent inference time, quantization discrepancy, and energy/power measurement methodology & Edge-processor benchmarking must report workload, hardware path, transfer overhead, and metric scope & Supports measurement discipline; does not benchmark SAT-Edge-Agent's FastAPI/LLM/YOLO workflow directly \\
SAT-Edge-Agent HIL evidence & Repeated fixed-workload YOLO-tool and full-Agent timing, CPU/NPU telemetry, structured partial-failure behavior, and metadata-backed artifacts & A mission-facing Agent workflow can coordinate local tools and return structured remote-sensing results on a COTS ARM-based heterogeneous edge SoC & Current evidence covers two fixed HIL workloads; it should not be generalized to broad mission distributions, flight performance, or calibrated energy efficiency \\
\bottomrule
\end{tabular}
\caption{Comparative positioning of SAT-Edge-Agent against measured onboard-AI and edge-processor studies. The table separates model-level performance evidence from workflow-level HIL evidence so that the paper does not imply an invalid speedup comparison.}
\label{tab:comparative_positioning}
\end{table}

This distinction constrains the efficiency claim. Because the tasks, processors, data units, and output contracts differ, the paper does not claim a universal speedup over prior onboard-AI systems. Its measured contribution is workflow-level: a YOLO-style OBB endpoint is embedded in a mission-facing path that also performs agent routing, annotated-result production, local response formation, SSE-visible state streaming, and structured artifact generation. The detector stage averaged 0.861 s for the single image and 1.511 s cumulatively for the serial pair, only about 2--3\% of the corresponding Full-Agent means. This separation identifies the broader orchestration and response path as the primary optimization target without assigning all remaining time to the language model. Payload-size and transfer-time effects of returning structured results instead of raw images are evaluated separately in the public downlink-payload experiment.

\section{System Architecture}
\label{sec:architecture}

\subsection{Architecture Overview}

SAT-Edge-Agent consists of four active runtime layers and one evidence layer: a browser-based operator workspace, a FastAPI agent backend, a project-internal YOLO-style OBB detector service, a local LLM service, and an artifact boundary that records the runtime and service state. Table~\ref{tab:system_layers} summarizes the division of responsibility.

\begin{table}[htbp]
\centering
\renewcommand{\arraystretch}{1.16}
\setlength{\tabcolsep}{4pt}
\caption{SAT-Edge-Agent system layers. The detector is a callable Agent tool, while the local LLM supplies reasoning and response formation; neither is presented as the paper's standalone algorithmic contribution.}
\label{tab:system_layers}
\begin{tabular}{p{0.22\linewidth}p{0.28\linewidth}p{0.40\linewidth}}
\toprule
\rowcolor{gray!12}
\textbf{Layer} & \textbf{Component} & \textbf{Role} \\
\midrule
Operator workspace & React/Vite frontend & Browser interaction and streamed agent result display. \\
\rowcolor{graybg}
Agent service & FastAPI with LangChain/LangGraph components & Prompt handling, tool invocation, state/checkpoint support, and SSE streaming. \\
Vision tool & YOLO-style OBB endpoint & Oriented remote-sensing detection and FAIR1M metadata-backed geographic target fields. \\
\rowcolor{graybg}
Reasoning service & Local OpenAI-compatible LLM endpoint & Local language-model response and mission-summary generation. \\
Evidence layer & Repository artifacts and sanitized runtime records & Reproducibility boundary, service contracts, release notes, and unsupported-claim guardrails. \\
\bottomrule
\end{tabular}
\end{table}

\begin{figure*}[t]
\centering
\includegraphics[width=0.98\textwidth]{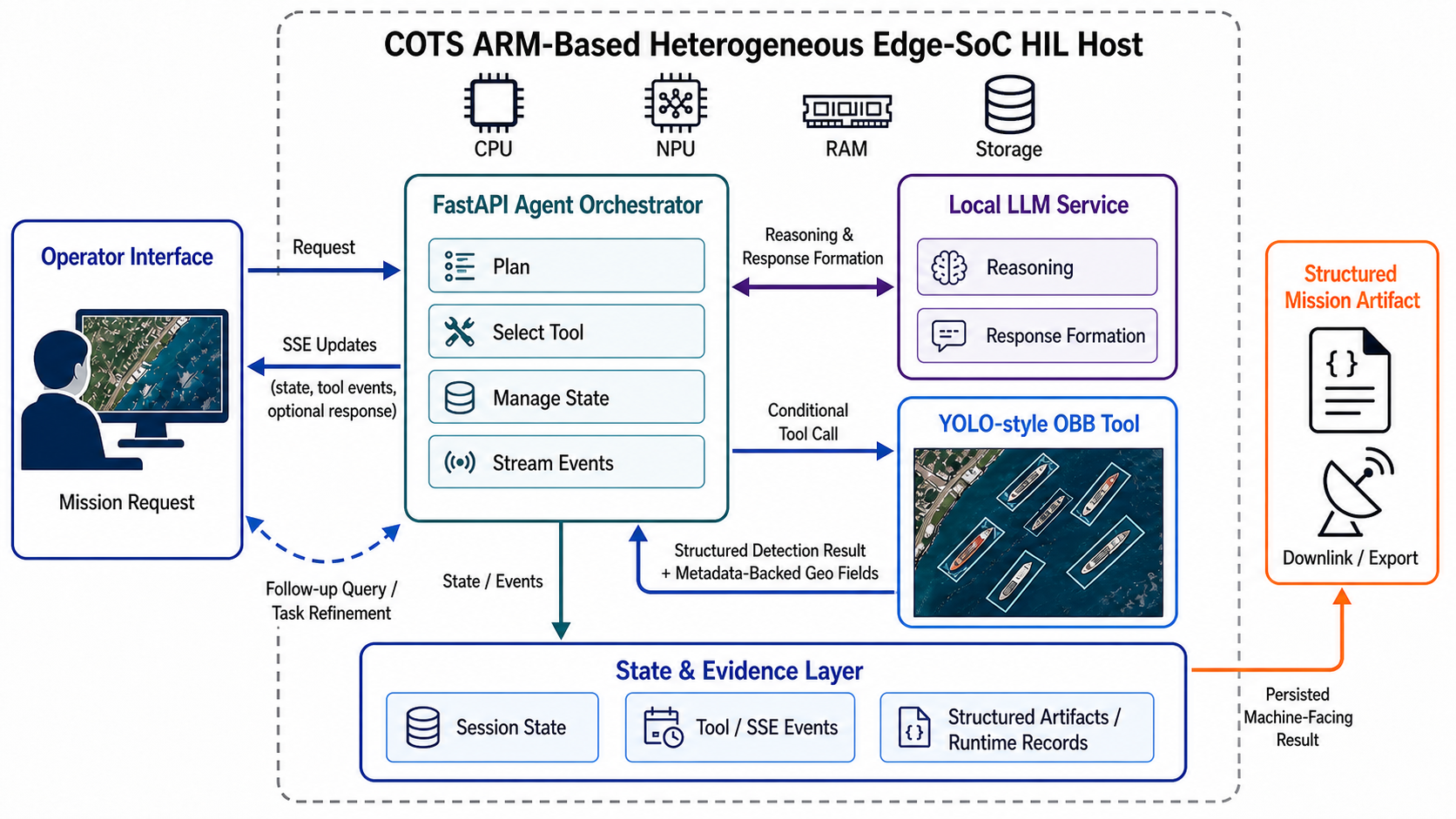}
\caption{SAT-Edge-Agent system overview. The FastAPI Agent uses the local LLM service for reasoning and response formation, conditionally invokes the YOLO-style OBB tool, and records tool events and structured results in the evidence layer. Machine-facing artifacts can be exported directly, while an operator-facing narrative is optional.}
\label{fig:architecture}
\end{figure*}

\subsection{Hardware-in-the-Loop Boundary}

The HIL boundary in this paper is the edge host and its local services. The recorded runtime snapshot identifies the host as an aarch64 Debian 11 environment running on a COTS ARM-based heterogeneous edge-SoC board. The public manuscript uses sanitized host labels such as \texttt{<edge-host>} and does not expose private IP addresses or access methods. The claim is therefore that SAT-Edge-Agent has been integrated and observed on a COTS ARM-based heterogeneous edge-SoC environment with local detection and language-model services. It is not a claim that the system has been flight-validated or qualified for the space radiation environment.

\subsection{End-to-End Execution Path}

The end-to-end path is intentionally compact. A mission request enters through the browser workspace, the FastAPI backend constructs agent state, and the backend invokes the detector when image-grounded evidence is required. The detector returns an authoritative structured artifact containing OBB detections and metadata-backed target fields. That artifact is emitted through the tool event for machine consumption before the local language service forms an optional mission-facing summary. SSEs expose the execution sequence to the operator instead of revealing only the final text. The workflow therefore converts input imagery into a structured mission product before downlink while keeping the narrative layer replaceable.

\subsection{Service Interfaces}

The detection service exposes \texttt{POST /v1/detect}, \texttt{GET /v1/health}, and \texttt{GET /v1/model/status}. The evidence package records a healthy endpoint response and a loaded project-internal OBB weight, image size 1024, object threshold 0.25, and non-maximum-suppression threshold 0.45. The private weight filename is withheld. In this manuscript, YOLO26 is a project-internal service label for a YOLO-style OBB endpoint, not a new public detector family or a standardized YOLO release. These records support service availability and configuration, not detector accuracy. The local LLM service is exposed through an OpenAI-compatible endpoint so that the orchestration layer can remain stable even when the underlying model implementation changes. Model-specific capability tests remain outside the public reproduction claim because their configurations and task boundaries differ from the repeated HIL workflow reported here.

\section{Agent Workflow and Tool Orchestration}
\label{sec:workflow}

\subsection{Workflow Concept}

The workflow begins with a user or mission request that references one or more remote-sensing images. The backend determines whether detection is required, invokes the vision tool, and receives structured detections. It then exposes the tool artifact and, when requested, produces a concise natural-language interpretation. The frontend receives both forms through SSEs.

The intended event lifecycle is \texttt{start} $\rightarrow$ \texttt{tool} $\rightarrow$ \texttt{token} $\rightarrow$ \texttt{done}, with structured error fields for failed tool calls or invalid input. The \texttt{tool} event is the machine-facing result boundary; the later \texttt{token} events serve the operator-facing narrative path. The current code and logs support SSE streaming and successful \texttt{/api/v1/chat/stream} calls. The public HIL artifact package includes normalized success and serial-batch partial-failure transcripts with request identifiers, runtime paths, base64 images, private labels, and operator text removed.

\begin{figure*}[t]
\centering
\includegraphics[width=0.98\textwidth]{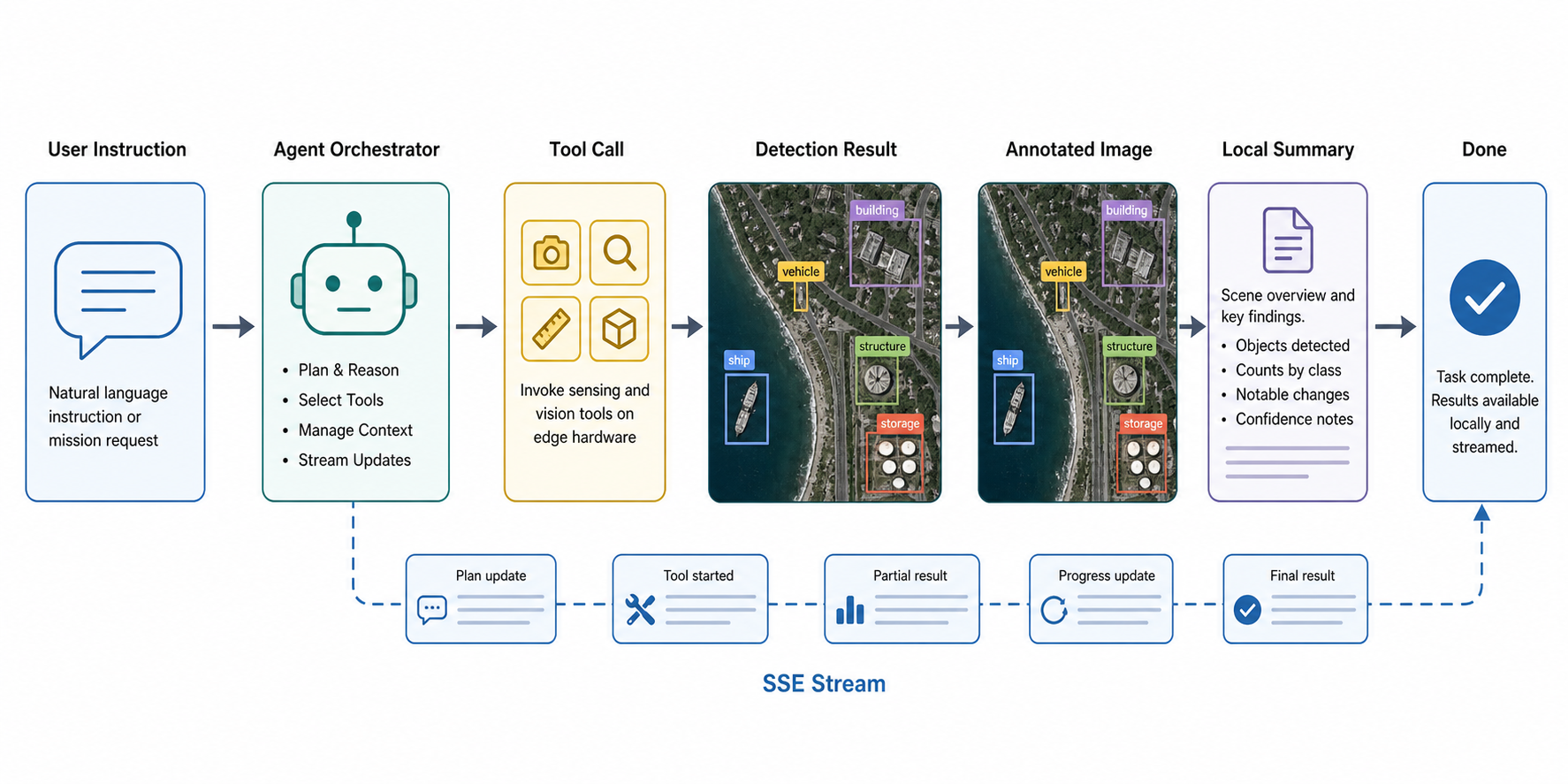}
\caption{HIL Agent workflow. The backend receives a user instruction, selects and invokes local tools, converts detections into annotated artifacts and a local summary, and streams intermediate events to the operator interface through SSE.}
\label{fig:workflow}
\end{figure*}

\subsection{Single-Image Analysis}

In the single-image workflow, the agent calls the detector once and receives a structured response containing image metadata, target detections, confidence scores, OBB polygons, pixel centers, and metadata-backed geographic centers. The agent then summarizes what was detected and whether geographic target fields are present. The representative case in the evidence package reports:
\begin{quote}
\texttt{detections=7; classes=A220:2, other-airplane:5; geo\_targets=7; total\_ms=846.646}
\end{quote}
This supports the claim that the system can produce metadata-backed structured mission artifacts from a FAIR1M-style image. The same fixed image is used by the repeated single-image timing protocol in Section~\ref{sec:results}; the representative response illustrates the output schema, whereas the 20-run experiment supports timing and repeatability claims. A compact redacted JSON record is shown with the service-contract evidence, and the complete sanitized example is included in the public artifact package.

\subsection{Failure Handling}

The evidence package includes a deliberately invalid upload. The endpoint returns a structured unsupported-media-type error with the message \texttt{Only jpg/png are supported}, and the service log includes a corresponding \texttt{415 Unsupported Media Type} record. A separate profiler-enabled serial batch exercised a missing second-image path. The tool returned \texttt{images\_count=2}, \texttt{success\_count=1}, and \texttt{failure\_count=1}, preserved a per-image error field, streamed a final response, and reached the SSE \texttt{done} event. Identifiers and runtime paths are omitted from the manuscript. These cases support structured invalid-input and partial-batch reporting, but they do not establish timeout recovery, retries, watchdog behavior, or service-restart correctness.

\subsection{Metadata-Backed Detection Artifacts}

The successful detection example includes request identifier, image file name, width, height, center geographic coordinates, per-target class ID and class name, confidence score, oriented polygon coordinates, pixel center, and geographic center. These geographic fields are not presented as a new geolocation algorithm. In the current evidence package, they are parsed from FAIR1M-style dataset metadata associated with the sample images and objects. When associated metadata are unavailable, the system treats geographic fields as unavailable rather than synthesizing coordinates; no sensor-model or map-registration fallback is implemented. The paper therefore claims metadata propagation and serialization into mission records, not independently validated absolute geolocation accuracy.

For \texttt{train\_\_t\_10144.jpg}, the detection output reports seven targets. Two are classified as \texttt{A220}, and five are classified as \texttt{other-airplane}. Each target contains a geographic center field, allowing the system to return metadata-backed target locations rather than only image-space boxes. This result is a qualitative system-output example. It is not an accuracy benchmark or a geolocation-error benchmark.

\section{Experimental Setup and Evidence}
\label{sec:results}

\subsection{Edge Runtime Snapshot}

Table~\ref{tab:runtime_snapshot} summarizes the HIL edge runtime snapshot used by the current manuscript. The table supports the setup description and should not be interpreted as a space-grade hardware qualification.

\begin{table}[htbp]
\centering
\renewcommand{\arraystretch}{1.16}
\setlength{\tabcolsep}{4pt}
\caption{HIL edge runtime snapshot recorded for the SAT-Edge-Agent evidence package. Private hostnames and paths are sanitized.}
\label{tab:runtime_snapshot}
\begin{tabular}{p{0.30\linewidth}p{0.55\linewidth}}
\toprule
\rowcolor{gray!12}
\textbf{Item} & \textbf{Evidence value} \\
\midrule
Operating system & Debian GNU/Linux 11 (bullseye) \\
\rowcolor{graybg}
Kernel/architecture & Linux 6.1.115, aarch64 \\
Board model & COTS ARM-based heterogeneous edge-SoC board (exact vendor model retained in internal evidence) \\
\rowcolor{graybg}
CPU and memory & 8 Cortex-A55 cores, max 2304 MHz; 31 GiB memory \\
Python runtime & Python 3.13.12 \\
\rowcolor{graybg}
Key packages & requests 2.32.5; FastAPI 0.136.1; Uvicorn 0.46.0; Pydantic 2.13.3; NumPy 2.4.4 \\
Runtime source state & Sanitized development snapshot; public reproduction release linked in \ref{app:artifacts} \\
\bottomrule
\end{tabular}
\end{table}

\subsection{Service-Contract Evidence}

The YOLO-style service evidence records a Uvicorn startup for \texttt{obb\_geo\_api\_server:app} with host \texttt{0.0.0.0} and port \texttt{8003}. The service was observed with a loaded project-internal OBB weight, image size 1024, object threshold 0.25, and NMS threshold 0.45. The exact private weight filename is retained only in internal evidence. Logs show multiple \texttt{POST /v1/detect 200 OK} entries and a deliberate \texttt{415 Unsupported Media Type} failure case. The label YOLO26 is used for this project service and should not be read as a new public detector family.

\begin{table}[htbp]
\centering
\renewcommand{\arraystretch}{1.16}
\setlength{\tabcolsep}{4pt}
\caption{Service-contract evidence for the YOLO OBB tool. These checks support API availability and structured behavior, not detector accuracy.}
\label{tab:service_evidence}
\begin{tabular}{p{0.30\linewidth}p{0.28\linewidth}p{0.30\linewidth}}
\toprule
\rowcolor{gray!12}
\textbf{Evidence item} & \textbf{Observed value} & \textbf{Claim supported} \\
\midrule
Health endpoint & \texttt{status=ok}, version 1.0.0 & Service availability \\
\rowcolor{graybg}
Model status & \texttt{loaded=true}, image size 1024 & Model-loaded service state \\
Success case & Seven detections with metadata-backed geographic fields & Structured output schema \\
\rowcolor{graybg}
Invalid upload & \texttt{UNSUPPORTED\_\allowbreak MEDIA\_\allowbreak TYPE}, HTTP 415 & Structured failure path \\
Partial serial batch & 1/2 images successful; final SSE \texttt{done} emitted & Structured partial-result path \\
\bottomrule
\end{tabular}
\end{table}

The public package preserves the exact field names while removing the request identifier and base64 image. The following compact record shows one of seven detections; coordinates are rounded for presentation.

\begin{small}
\begin{verbatim}
{
  "image": {"file_name": "train__t_10144.jpg",
            "width": 1000, "height": 1000},
  "detections_count": 7,
  "detections": [{
    "class_name": "A220", "confidence": 0.968452,
    "polygon": [[598.936,136.701],[619.303,91.140],
                [581.040,74.035],[560.673,119.596]],
    "pixel_center": [589.988,105.368],
    "geo_center": [118.121802,24.539276]
  }],
  "perf": {"total_ms": 846.646}, "geo_status": "ok"
}
\end{verbatim}
\end{small}

\subsection{Metadata-Backed Output Evidence}

The representative FAIR1M case returns seven detections with metadata-backed geographic target fields. The output includes aircraft classes, confidence values, OBB polygons, pixel centers, and geographic centers parsed from FAIR1M-style sample metadata. This supports the claim that the system can serialize detection outputs into mission-facing records. It does not support an accuracy benchmark, SOTA detector claim, or independently validated geolocation-error claim.

\subsection{Repeated Fixed-Workload HIL Protocol}

The primary performance evidence was collected on 2026-07-13 using two fixed FAIR1M workloads. The single-image workload repeatedly uses \texttt{train\_\_t\_10144.jpg}; the serial workload uses that image followed by \texttt{train\_\_t\_10175.jpg}. Warm-up requests were excluded, and the two-image workload invokes the images serially rather than concurrently. Each workload contains 20 measured attempts. Full-Agent latency starts when the Agent receives \texttt{/api/v1/chat/stream} and ends when the SSE stream emits \texttt{done}; it includes Agent routing, detector invocation, result handling, annotated-output production, local response formation, and SSE delivery. YOLO-tool latency is the detector-stage duration reported by the vision tool. CPU and NPU fields were sampled over the full-Agent window.

The resource sampler queried CPU counters and a devfreq/sysfs NPU load/frequency field at 200-ms intervals. The NPU field spans the complete request and is shared by accelerator consumers, including the detector and local language service. Consequently, the reported 100\% value is a coarse shared-accelerator software field, not detector-only occupancy, per-service attribution, or calibrated utilization. It measures a different quantity from the detector's wall-clock share of Full-Agent latency.

\begin{table}[htbp]
\centering
\small
\renewcommand{\arraystretch}{1.14}
\setlength{\tabcolsep}{3.5pt}
\caption{Fixed-workload protocol for the repeated HIL experiment. The sample counts characterize repeatability for these two workloads, not a FAIR1M-wide latency distribution.}
\label{tab:fixed_workload_protocol}
\begin{tabular}{p{0.23\linewidth}p{0.30\linewidth}p{0.17\linewidth}p{0.18\linewidth}}
\toprule
\rowcolor{gray!12}
\textbf{Scenario} & \textbf{Fixed input} & \textbf{Output per run} & \textbf{Attempts} \\
\midrule
Single image & \texttt{train\_\_t\_10144.jpg} & 7 detections & 20; 20/20 completed \\
\rowcolor{graybg}
Serial two-image & \texttt{train\_\_t\_10144.jpg}\newline + \texttt{train\_\_t\_10175.jpg} & 18 detections & 20; 20/20 completed \\
\bottomrule
\end{tabular}
\end{table}

For the statistics below, sample standard deviation uses $n-1$ in the denominator, the median is the ordinary sample median, and empirical P95 is the nearest-rank order statistic $x_{\lceil0.95n\rceil}$. P99 is omitted because at $n=20$ the nearest-rank estimate collapses to the sample maximum and does not support a robust tail claim.

\begin{figure*}[t]
\centering
\includegraphics[width=0.98\textwidth]{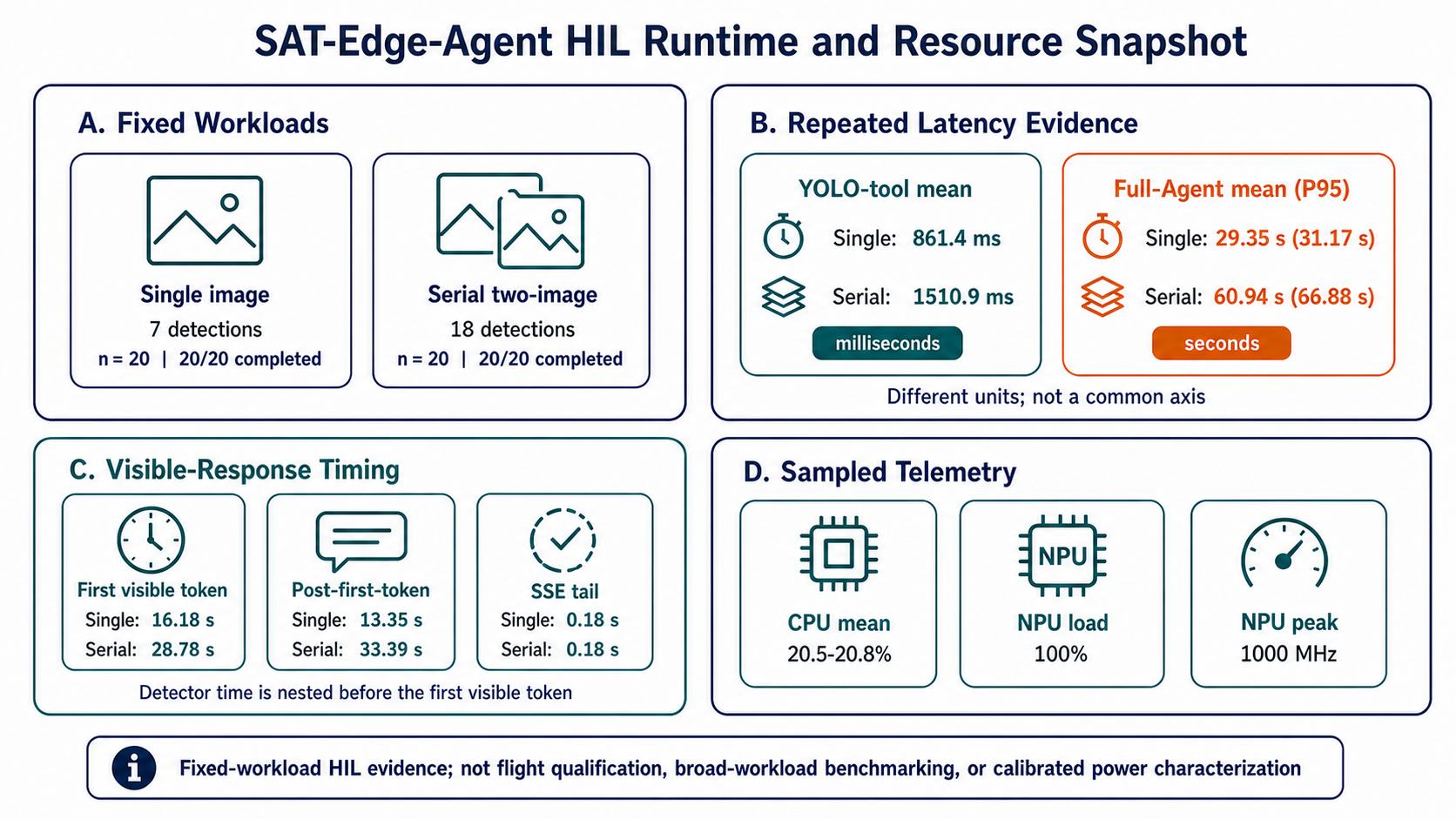}
\caption{Repeated HIL evidence for the two fixed FAIR1M workloads. Panel A defines the protocol. Panel B separates mean YOLO-tool latency in milliseconds from mean and empirical-P95 full-Agent latency in seconds; the cards do not encode a common axis. Panel C reports validated profiler intervals in seconds, with detector time identified as nested before the first visible response token rather than as an additive stage. Panel D reports the 200-ms sampled shared-accelerator NPU-load field and the $20.5$--$20.8\%$ range of the two per-workload mean CPU values. The separate 2026-07-03 plug-meter pilot is intentionally excluded.}
\label{fig:runtime_resource_snapshot}
\end{figure*}

\begin{figure*}[t]
\centering
\includegraphics[width=0.98\textwidth]{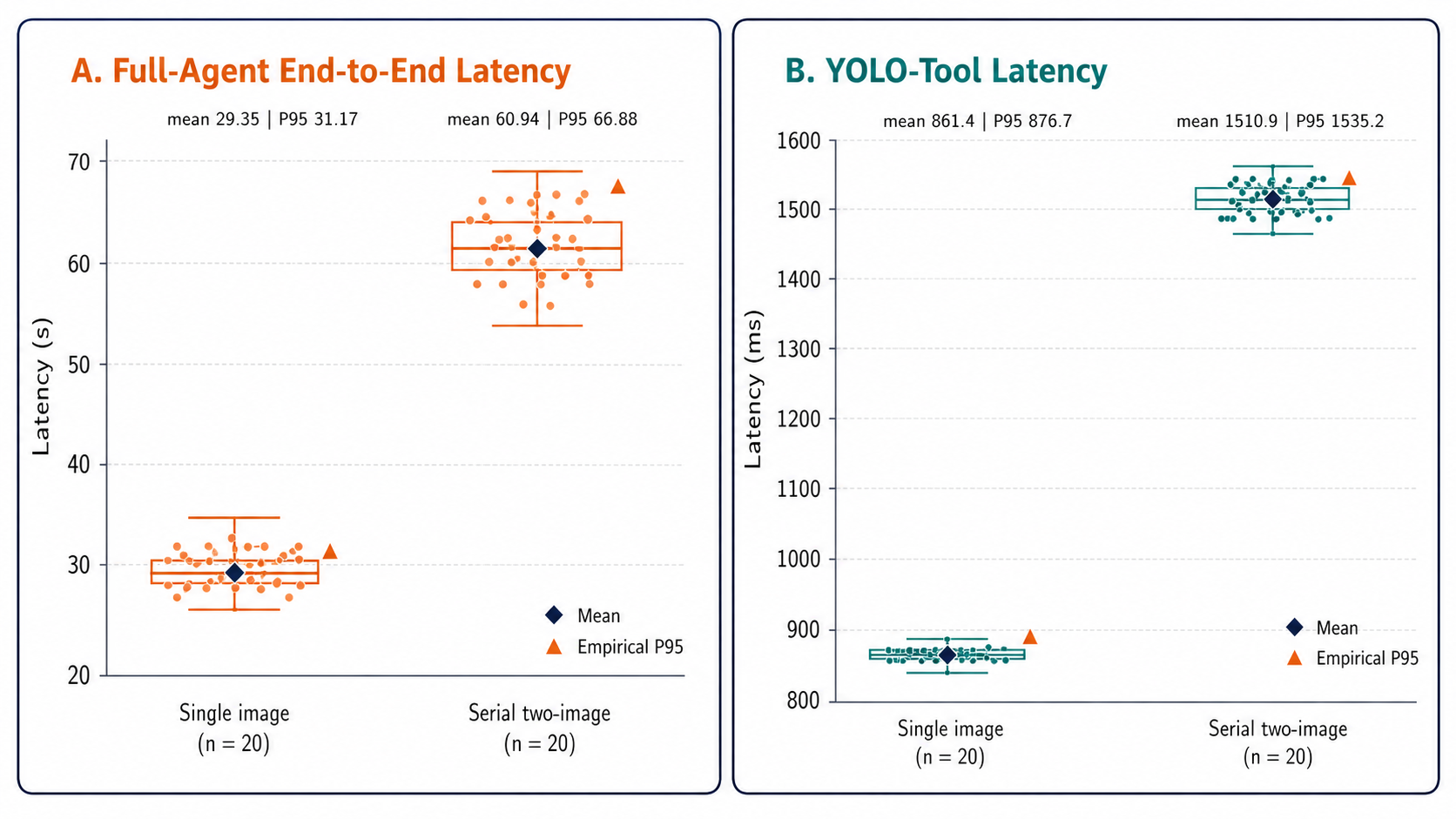}
\caption{Request-level latency distributions supplementing Figure~\ref{fig:runtime_resource_snapshot} for the two fixed FAIR1M workloads ($n=20$ each). Points denote individual completed requests, boxes show the interquartile range and median, diamonds mark means, and triangles mark empirical nearest-rank P95. Panel A measures Full-Agent latency from request receipt through SSE completion in seconds; Panel B reports the nested YOLO-tool duration in milliseconds. The panels use independent axes and do not constitute a cross-hardware benchmark or a FAIR1M-wide latency characterization.}
\label{fig:latency_distributions}
\end{figure*}

Figure~\ref{fig:latency_distributions} supplements the aggregate latency cards in Figure~\ref{fig:runtime_resource_snapshot} with the underlying request-level distributions for the same fixed workloads.

\begin{table}[htbp]
\centering
\small
\renewcommand{\arraystretch}{1.12}
\setlength{\tabcolsep}{4pt}
\caption{Repeated full-Agent, YOLO-tool, and sampled resource statistics for the fixed FAIR1M workloads ($n=20$ each). Resource fields were sampled every 200 ms across the full request. CPU peak is the maximum observed per-request peak field, not a platform limit; the NPU field is not detector-only occupancy.}
\label{tab:agent_runtime_resource}
\begin{tabular}{p{0.43\linewidth}p{0.23\linewidth}p{0.23\linewidth}}
\toprule
\rowcolor{gray!12}
\textbf{Metric} & \textbf{Single image} & \textbf{Serial two-image} \\
\midrule
Observed completion & 20/20 & 20/20 \\
\rowcolor{graybg}
Full-Agent mean $\pm$ sample SD & $29.353 \pm 1.613$ s & $60.937 \pm 3.512$ s \\
Full-Agent median & 29.030 s & 61.134 s \\
\rowcolor{graybg}
Full-Agent empirical P95 & 31.166 s & 66.882 s \\
Full-Agent range & 26.919--34.504 s & 54.379--68.263 s \\
\rowcolor{graybg}
YOLO-tool mean $\pm$ sample SD & $861.386 \pm 11.778$ ms & $1510.920 \pm 18.180$ ms \\
YOLO-tool median & 863.415 ms & 1513.604 ms \\
\rowcolor{graybg}
YOLO-tool empirical P95 & 876.657 ms & 1535.151 ms \\
Detector share of mean Full-Agent latency & 2.93\% & 2.48\% \\
\rowcolor{graybg}
Mean sampled CPU utilization & 20.761\% & 20.482\% \\
Maximum observed CPU peak field & 86.335\% & 87.742\% \\
\rowcolor{graybg}
Mean 200-ms sampled NPU-load field & 100\% & 100\% \\
Peak sampled NPU frequency & 1000 MHz & 1000 MHz \\
\bottomrule
\end{tabular}
\end{table}

For these fixed workloads, detector execution is a small fraction of the end-to-end request: 2.93\% of the single-image mean and 2.48\% of the serial two-image mean. The remaining interval cannot be assigned entirely to language-model generation because Full-Agent latency also includes agent decisions, request construction, tool-result handling, annotated-output production, and response streaming. The 100\% NPU field likewise does not imply continuous detector execution; it is a shared-accelerator software sample over the complete request.

\subsection{Validated Visible-Response Timeline}

A separate profiler-enabled run set captures the event order \texttt{start} $\rightarrow$ \texttt{tool} $\rightarrow$ \texttt{token} $\rightarrow$ \texttt{done}. All 20 single-image profiler attempts completed. Nineteen serial two-image attempts completed both images; one additional attempt produced the structured 1/2 partial result described in the workflow section and is excluded from the all-images-successful timing mean. Table~\ref{tab:response_timeline} uses only fields validated against the raw SSE captures.

\begin{table}[htbp]
\centering
\small
\renewcommand{\arraystretch}{1.13}
\setlength{\tabcolsep}{4pt}
\caption{Mean response-timeline intervals for successful profiler-enabled runs. ``First visible token'' is an end-to-end user-visible boundary that includes pre-output Agent and tool activity; it is not pure LLM first-token latency. Detector time is nested inside that interval and must not be added to the other rows.}
\label{tab:response_timeline}
\begin{tabular}{p{0.43\linewidth}p{0.23\linewidth}p{0.23\linewidth}}
\toprule
\rowcolor{gray!12}
\textbf{Metric} & \textbf{Single image} & \textbf{Serial two-image} \\
\midrule
All-images-successful attempts & 20 & 19 \\
\rowcolor{graybg}
Full-Agent mean & 29.713 s & 62.362 s \\
Time to first visible response token & 16.184 s & 28.785 s \\
\rowcolor{graybg}
Post-first-token generation interval & 13.347 s & 33.392 s \\
SSE tail after final token & 0.182 s & 0.185 s \\
\rowcolor{graybg}
Detector-total mean (nested) & 0.857 s & 1.506 s \\
\bottomrule
\end{tabular}
\end{table}

The component CSV also contains batch HTTP, decode, inference, and annotation/render substage fields. Audit against the raw serial-batch captures found a systematic factor-of-two aggregation defect, so those fields are excluded from the manuscript. The validated intervals support a narrower conclusion: most user-facing latency lies outside detector execution, both before and after the first visible token. Finer attribution among orchestration, rendering, and language-service substages requires corrected instrumentation.

\subsection{End-to-End Reproducibility Matrix}

Table~\ref{tab:repro_matrix} separates what a reader can reproduce directly from what remains private or replaceable. This distinction is important because the detector weight and private edge host are not the scientific contribution of the present paper.

\begin{table}[htbp]
\centering
\renewcommand{\arraystretch}{1.16}
\setlength{\tabcolsep}{4pt}
\caption{Reproducibility boundary for the SAT-Edge-Agent HIL system paper.}
\label{tab:repro_matrix}
\begin{tabular}{p{0.28\linewidth}p{0.28\linewidth}p{0.30\linewidth}}
\toprule
\rowcolor{gray!12}
\textbf{Layer} & \textbf{Publicly reproducible} & \textbf{Boundary or substitute path} \\
\midrule
Frontend/backend code & Build and run workflow, API routes, SSE stream contract & Private hostnames and paths are sanitized \\
\rowcolor{graybg}
Vision-tool contract & Health, model-status schema, success/failure JSON examples & Private YOLO weight can be replaced by a user-provided OBB model \\
Local LLM interface & OpenAI-compatible endpoint contract & Exact local model may be deployment-specific \\
\rowcolor{graybg}
Dataset sample & FAIR1M-style schema and license notes & FAIR1M-derived images must follow original dataset terms \\
Repeated timing evidence & Sanitized request-level CSVs, summary JSON/Markdown, recalculation script, and Figure 3 & Covers only the two declared fixed workloads \\
\rowcolor{graybg}
SSE and JSON contracts & Redacted success/failure JSON and normalized success/partial-result SSE examples & Raw identifiers, paths, base64 imagery, private labels, and operator text are omitted \\
Power evidence & Appendix-level plug-meter context and a public trace template & A synchronized calibrated trace is required before energy claims \\
\bottomrule
\end{tabular}
\end{table}

The reproducibility matrix separates public artifacts from replaceable or private components. The released package is located at \texttt{artifacts/hil\_orchestration/} and includes a SHA-256 manifest. Remaining workload-generalization, tail-latency, power, robustness, and metadata limitations are consolidated in the Limitations section and \ref{app:artifacts} rather than repeated as a second evidence-gap table.

\section{Discussion}

\subsection{System Value of the Agent Layer}

The system contribution is the contract between mission intent, local tools, and two output forms. The structured detector artifact is the authoritative machine-facing result. Natural-language synthesis is an optional operator-facing layer that can be shortened, deferred, or replaced without changing the detector contract. In the measured narrative path, the mean times to the first visible response token were 16.184 s and 28.785 s. These delays are unsuitable for time-critical closed-loop control. A time-sensitive machine service should instead consume the structured \texttt{tool} event and treat language synthesis as a separable operator aid. Because the current study does not include a schema-only versus narrative-enabled ablation, this architectural option is not presented as a measured latency reduction.

\subsection{HIL and Measurement Boundary}

The HIL deployment grounds the architecture in an observed edge-host runtime. The repeated experiment provides 20 completed attempts for each fixed workload, a validated visible-response timeline, repeated CPU/NPU telemetry, and a structured 1/2 partial-batch case. The NPU and latency results describe different quantities: the 100\% value is a 200-ms devfreq/sysfs software field sampled across a request in which both local inference services may use the accelerator, whereas the 2--3\% value is the detector's wall-clock share of Full-Agent latency. The NPU field therefore cannot be interpreted as detector occupancy or per-service utilization. The separate plug-meter pilot is not synchronized with these runs and remains Appendix-level context. Together, the measurements support bounded repeatability for the two workloads, not flight readiness. Radiation-aware design review, watchdog and recovery tests, environmental testing, and a flight- or RF-in-the-loop campaign remain necessary before an in-orbit claim is appropriate.

\subsection{Artifact and Release Boundary}

The public release provides sanitized request-level CSV files, redacted success and invalid-input JSON, normalized success and partial-failure SSE examples, a statistics script, and a SHA-256 manifest. The detector weight remains an internal training asset, and the exact board and local language-model identities are withheld. YOLO26 denotes only the project-internal YOLO-style OBB service. Readers can substitute compatible detector and language endpoints while preserving the documented contracts. FAIR1M-derived images remain governed by their source terms and are not redistributed in the public evidence package.

\section{Limitations}

Four boundaries constrain interpretation of the results. \textbf{Deployment scope.} The work reports a ground-based HIL edge deployment, not an in-orbit demonstration, radiation qualification, or flight-computer certification. \textbf{Perception and metadata.} YOLO26 is a project-internal label for a YOLO-style OBB tool; the paper does not report detector mAP, SOTA accuracy, a training-protocol comparison, or independently validated geolocation error. Geographic fields are propagated from FAIR1M-style metadata rather than derived from a validated sensor model.

\textbf{Measurement scope.} The repeated evidence covers one fixed FAIR1M image and one fixed serial two-image pair, with 20 attempts per workload; it does not establish performance over a broader distribution of images, target counts, prompts, batch sizes, or concurrency. Mean, sample SD, median, and empirical nearest-rank P95 are reported, but $n=20$ does not support robust P99 or general tail-latency claims. The first-visible-token interval includes agent and tool activity and is not pure language-model first-token latency; frontend rendering and several internal substages remain unisolated. Power values come from a separate plug-meter pilot, so a quantitative energy study still requires calibrated, synchronized traces and instrument metadata.

\textbf{Operational robustness and reproducibility.} The evidence demonstrates a structured invalid-file response and one serial-batch 1/2 partial result that reached \texttt{done}, but not comprehensive timeout, retry, watchdog, restart, or fault-recovery behavior. The public package exposes sanitized evidence, endpoint contracts, a vendor-neutral runtime profile, and recalculation scripts. Withholding the exact board identity, private language-model identity, detector weight, and raw logs limits hardware-normalized performance comparison, although compatible endpoints can reproduce the workflow contract. FAIR1M-derived images remain governed by their original dataset license.

\section{Conclusion and Future Work}

SAT-Edge-Agent demonstrates an evidence-bounded HIL implementation of satellite edge-agent orchestration on a COTS ARM-based heterogeneous edge SoC. The system connects a browser workspace, FastAPI agent backend, local language service, and YOLO-style OBB endpoint through an observable workflow. Both fixed FAIR1M workloads completed 20/20 attempts. Mean Full-Agent latency was 29.353 s for the single image and 60.937 s for the serial pair, while detector execution represented only 2.93\% and 2.48\% of those means. This separation is the central system-level result: for the measured workloads, the primary optimization opportunity lies in orchestration and response formation rather than detector execution alone. The evidence also verifies metadata-backed structured output, repeated CPU/NPU telemetry, visible-response timing, invalid-file handling, and a structured 1/2 partial-batch result.

Future work should broaden the workload distribution, evaluate concurrency and larger image batches, isolate finer orchestration and frontend stages, and collect enough runs for defensible tail-latency analysis. A controlled schema-only versus narrative-enabled experiment is required before quantifying the benefit of deferring natural-language synthesis. Calibrated power and energy measurements, watchdog and fault-injection tests, and an RF or flight-like environmental campaign remain separate requirements for a more complete onboard edge-agent systems study.

Code and sanitized research artifacts are available at \url{https://github.com/keithhegit/SAT-Edge-Agent}. The release documents the public evidence, dataset-license notes, model/runtime boundary, and reproduction procedure. Private deployment details remain replaced by placeholders such as \texttt{<edge-host>}; \ref{app:artifacts} lists the public inventory and release boundary.

\clearpage
\appendix
\renewcommand{\thesection}{Appendix \Alph{section}}

\section{Reproducibility and Release Boundary}
\label{app:artifacts}

This appendix records the public evidence, measurement definitions, and release boundary used by the current SAT-Edge-Agent manuscript. The public release is available at \url{https://github.com/keithhegit/SAT-Edge-Agent}; all paths below are relative to that repository root.

\subsection{Public Artifact Inventory}

\begin{table}[H]
\centering
\footnotesize
\renewcommand{\arraystretch}{1.15}
\setlength{\tabcolsep}{4pt}
\caption{Public artifacts supporting the HIL orchestration claims.}
\label{tab:hil_artifacts}
\begin{tabular}{p{0.34\linewidth}p{0.53\linewidth}}
\toprule
\rowcolor{gray!12}
\textbf{Artifact} & \textbf{Repository path} \\
\midrule
Public evidence guide & \texttt{artifacts/hil\_orchestration/README.md} \\
\rowcolor{graybg}
Repeated fixed-workload rows & \texttt{artifacts/hil\_orchestration/data/}
\newline\texttt{fixed\_workload\_runs.csv} \\
Visible-response profiler rows & \texttt{artifacts/hil\_orchestration/data/}
\newline\texttt{visible\_response\_timing.csv} \\
\rowcolor{graybg}
Derived result summaries & \texttt{artifacts/hil\_orchestration/results/}
\newline\texttt{hil\_orchestration\_summary.json}, \texttt{hil\_orchestration\_summary.md} \\
Recalculation script & \texttt{artifacts/hil\_orchestration/scripts/}
\newline\texttt{summarize\_public\_metrics.py} \\
\rowcolor{graybg}
Redacted API examples & \texttt{artifacts/hil\_orchestration/examples/}
\newline\texttt{detect\_success\_redacted.json}, \texttt{detect\_failure\_redacted.json} \\
Normalized SSE examples & \texttt{artifacts/hil\_orchestration/examples/}
\newline\texttt{sse\_success\_normalized.jsonl},
\newline\texttt{sse\_partial\_failure\_normalized.jsonl} \\
\rowcolor{graybg}
Integrity manifest & \texttt{artifacts/hil\_orchestration/MANIFEST.sha256} \\
Model/runtime and reproduction boundary & \texttt{MODEL\_CARD.md}, \texttt{REPRODUCIBILITY.md} \\
\rowcolor{graybg}
Dataset acquisition and checksums & \texttt{dataset/README.md}, \texttt{dataset/DATA\_LICENSE.md},
\newline\texttt{dataset/sample\_100\_mix\_manifest.csv},
\newline\texttt{dataset/sample\_100\_mix\_sha256.csv} \\
Power-trace field template & \texttt{artifacts/hil\_orchestration/templates/}
\newline\texttt{power\_trace\_template.csv} \\
\bottomrule
\end{tabular}
\end{table}

The public CSV rows are field-level redactions of the internal request records. They remove source-capture names, timestamps, request and thread identifiers, input paths, exact private model labels, and operator text. Running the public script recalculates the means, sample standard deviations, medians, nearest-rank empirical P95 values, detector shares, and profiler means reported in the manuscript.

\subsection{Normalized Service Examples}

The normalized success transcript preserves the event order and selected validated values while omitting payload imagery and natural-language content:

\begin{small}
\begin{verbatim}
event: start
event: tool  {"phase":"result","success_count":1,
              "failure_count":0,"total_detections_count":7}
event: token {"text":"<operator-facing summary omitted>"}
event: done  {"duration_ms":29722,
              "detector_total_ms":851.794,
              "time_to_first_visible_token_ms":19966}
\end{verbatim}
\end{small}

The partial-result example records two requested inputs, one successful input, one unavailable input, seven retained detections, a final operator-facing response, and the terminal \texttt{done} event:

\begin{small}
\begin{verbatim}
event: start
event: tool  {"phase":"result","success_count":1,
              "failure_count":1,"total_detections_count":7}
event: token {"text":"<partial-result summary omitted>"}
event: done  {"duration_ms":57300,
              "completion_status":"partial_result"}
\end{verbatim}
\end{small}

These examples support event-contract and structured-degradation claims. They do not establish retry, timeout recovery, watchdog behavior, or service restart correctness.

\subsection{Telemetry and Power Boundaries}

CPU counters and the NPU devfreq/sysfs load/frequency field were sampled every 200 ms across the complete Full-Agent request. The detector and local language service can both use the accelerator. The reported 100\% mean NPU-load field is therefore a coarse shared-accelerator software value rather than detector-only occupancy, per-service attribution, or calibrated utilization.

Power observations were collected in a separate 2026-07-03 pilot with different workloads. They are retained as board-level HIL context but are not synchronized with the repeated FAIR1M measurements. The manuscript therefore does not derive energy per request by multiplying the following observations by the 2026-07-13 latency statistics.

\begin{table}[H]
\centering
\renewcommand{\arraystretch}{1.14}
\setlength{\tabcolsep}{4pt}
\caption{Board-level plug-meter observations from the separate pilot. These values do not constitute a calibrated continuous spacecraft power trace.}
\label{tab:power_snapshot}
\begin{tabular}{p{0.24\linewidth}p{0.18\linewidth}p{0.43\linewidth}}
\toprule
\rowcolor{gray!12}
\textbf{Operating state} & \textbf{Observed power} & \textbf{Evidence boundary} \\
\midrule
Idle baseline & $\sim$3 W & Manual plug-meter screen observation before request and after recovery \\
\rowcolor{graybg}
Active-load interval & $\sim$8--9 W & Manual observation during detection and local Agent response formation \\
Post-output transition & $\sim$4 W & Manual observation after output and before idle recovery \\
\bottomrule
\end{tabular}
\end{table}

The public trace template declares the fields needed for a stronger study: timestamp, workload state, voltage, current, power, board temperature, ambient temperature, and note. A calibrated instrument description, measured rail, sampling period, timestamp alignment, and synchronized workload script remain necessary before reporting peak power, energy per request, or thermal behavior.

\subsection{Data, Model, and Privacy Boundary}

FAIR1M-derived images, labels, and geographic fields are third-party data governed by the original FAIR1M/Kaggle terms. The repository publishes source attribution, download instructions, a 100-file manifest, image checksums, and a narrowly redacted output example, but not the image files or full internal metadata bundle. FAIR1M-derived fields in the example remain subject to the source data terms rather than the repository software license. The detector weight is an internal training asset and is not redistributed. The exact board model, exact private language-model identity, private generation throughput, hostnames, IP addresses, VM access methods, internal paths, and raw unredacted logs are also withheld.

The public reproduction path is contract-based. Readers may use a compatible OBB endpoint and an OpenAI-compatible local language endpoint, or recalculate the manuscript statistics directly from the released sanitized evidence. The earlier Qwen Instruct 16K capability benchmark remains separate supplementary material and is not identified as the exact configuration used by the 2026-07-13 repeated HIL experiment.

\subsection{Claim Boundary}

The current manuscript does not claim flight validation, radiation qualification, detector SOTA accuracy, complete model-weight reproduction, independently validated geolocation error, broad-workload latency generalization, robust P99 latency, calibrated power or energy characterization, or a complete autonomous mission scheduler. Each stronger claim would require evidence beyond the fixed-workload HIL measurements, public service examples, runtime profile, and separate plug-meter pilot reported here.

\end{document}